\documentclass[a4paper,conference]{IEEEtran}
\IEEEoverridecommandlockouts
\usepackage{cite}
\usepackage{amsmath,amssymb,amsfonts}
\usepackage{algorithmic}
\usepackage{graphicx}
\usepackage{textcomp}
\usepackage{tikz}
\usepackage{pgfplots}
\usetikzlibrary{pgfplots.groupplots}
\usetikzlibrary{positioning}
\usetikzlibrary{arrows}
\usetikzlibrary{calc,patterns,angles,quotes}
\usepackage{tkz-euclide}
\usepackage{tikzscale}
\usepackage{xcolor}
\def\BibTeX{{\rm B\kern-.05em{\sc i\kern-.025em b}\kern-.08em
    T\kern-.1667em\lower.7ex\hbox{E}\kern-.125emX}}
\begin{document}

\title{Wasserstein Routed Capsule Networks
}

\author{\IEEEauthorblockN{Alexander Fuchs}
\IEEEauthorblockA{\textit{SPSC} \\
\textit{Graz University of Technology}\\
Graz, Austria\\
fuchs@tugraz.at}
\and
\IEEEauthorblockN{Franz Pernkopf}
\IEEEauthorblockA{\textit{SPSC} \\
\textit{Graz University of Technology}\\
Graz, Austria \\
pernkopf@tugraz.at}
}

\maketitle

\begin{abstract}
Capsule networks offer interesting properties and provide an alternative to today's deep neural network architectures. However, recent approaches have failed to consistently achieve competitive results across different image datasets. We propose a new parameter efficient capsule architecture, that is able to tackle complex tasks by using neural networks trained with an approximate Wasserstein objective to dynamically select capsules throughout the entire architecture. This approach focuses on implementing a robust routing scheme, which can deliver improved results using little overhead.
We perform several ablation studies verifying the proposed concepts and show that our network is able to substantially outperform other capsule approaches by over 1.2 $\%$ on CIFAR-10, using fewer parameters. 
\end{abstract}


\section{Introduction}
Todays computer vision systems mostly rely on large deep neural networks (DNNs). Sophisticated methods have been proposed to train structures hundreds of layers deep, achieving superhuman performance on speech and image processing tasks \cite{he2016deep,huang2016deep,DBLP:journals/corr/HuangLW16a}. All of today's DNN architectures use convolutional layers (CNNs) \cite{Cun:1990:HDR:109230.109279}, which have the advantage of local connectivity due to the filter kernels being shifted over the image, implementing a translational invariance of features with respect to the feature positions. However, the networks still need to learn different filters for various object orientations and sizes, which also means that all of these variations need to be included in the dataset. This issue is often tackled by using data augmentation techniques such as, rotating, flipping and resizing the image. Since most of the objects in image datasets are 2D projections of 3D objects, data augmentation is limited to a small set of possible augmentations if no 3D model of the underlying object is available. Capsule networks (CapsNets) try to solve this by learning equivariant representations on a part or object level, i.e. the networks try to learn an object representation independent of its orientation and size \cite{Hinton2011,kosiorek2019stacked}.\newline
CapsNets fundamentally rely on routing schemes to select and combine different capsules for classification. These routing schemes assess a capsule according to a pre-defined criterion and assign a weighting factor to each capsule to indicate the strength of its presence in the routing result. This principle allows for the specialization of capsules, but also introduces the problem of incorrect routings,leading to wrong classification results. Recent CapsNet approaches perform well on simple datasets, where the objects are clearly separable from the background, but have difficulties if the images also contain background information \cite{NIPS2017_6975,xi2017capsule}. To a certain extent, this can be solved by using a DNN as a pre-processing stage for the CapsNet \cite{DBLP:journals/corr/abs-1904-09546,NIPS2018_7823}. Unfortunately, CapsNets still fail to achieve competitive results for large and complex datasets, partly due to the bad scalability of the capsule architecture to many classes. Therefore, fundamental changes in the used architectures need to be introduced to make CapsNets applicable to a larger set of problems.\newline
In this paper, we propose a new Wasserstein Capsule Network architecture (WCapsNet), which focuses on efficiency and scalability, making CapsNets applicable to a wide class of computer vision problems. We propose an architecture that uses a critic CNN trained with a Wasserstein objective to solve the problem of capsule routing. This routing joins the multiple levels of the WCapsNet architecture \cite{pmlr-v70-arjovsky17a}, and enables the specialization of the feature detectors across multiple abstraction levels. To train the critic networks, we propose an approximation scheme for the Wasserstein objective, suitable for capsule routing. This highly dynamic WCapsNet architecture implements a parameter efficient classification network. Furthermore, we introduce a vector non-linearity suitable for the WCapsNet architecture. The non-linearity acts on the direction of the capsule vectors and tilts them toward strong components. To validate the proposed Wasserstein routing and the vector non-linearity, we perform several ablation studies presented in Section \ref{sec:ablation_studies}.
Our proposed WCapsNet architecture offers an efficient and scale-able approach for image classification and improves the interpretability of DNNs, by offering possibilities to identify the most relevant parts of the networks for specific input classes. We substantially outperform other capsule approaches by over 1.2 $\%$ on CIFAR-10, and show that the architecture is able to deliver a good performance for a more complex dataset like CIFAR-100, without having large computational overhead.
 \section{Related work}
The first capsule architecture used for classification \cite{NIPS2017_6975} works well for relatively simple datasets, but fails to achieve competitive results for more complex data \cite{xi2017capsule}.
Improvements in terms of classification performance have been achieved by using additional DNN architectures as a pre-processing stage for the CapsNets \cite{DBLP:journals/corr/abs-1904-09546,NIPS2018_7823}. Several papers proposed improvements to the routing, using unsupervised routing-algorithms, but failed to consistently achieve good performances across datasets \cite{improved_routing}. Recently, other approaches for solving the dynamic routing problem have been proposed. In particular, supervised methods, such as neural networks, are used for an improved weight assignment \cite{hahn2019self}, or to generate attention maps which are combined with a binary gating function, trained with the Straight-Through estimator \cite{ahmed2019star,bengio2013estimating}. Less classification focused papers have shown the usefulness of using capsules as parts for object reconstruction in 2D and also for 3D point clouds. With stacked capsule-autoencoders achieving state-of-the-art results for unsupervised classification \cite{kosiorek2019stacked,Zhao_2019_CVPR}. A different approach of finding equivariant representations is to explicitly include the invariances in the convolutions \cite{cohen2016group}. This approach generalizes the translation equivariance of standard convolutions used in computer vision, to convolutions invariant with respect to any transformation from a specific symmetry group, leading to equivariance on a feature, rather than a part or object level. 

\section{Wasserstein Capsules Network (WCapsNet)}
\begin{figure*}
    \begin{center}
    \includegraphics[width=0.7\linewidth]{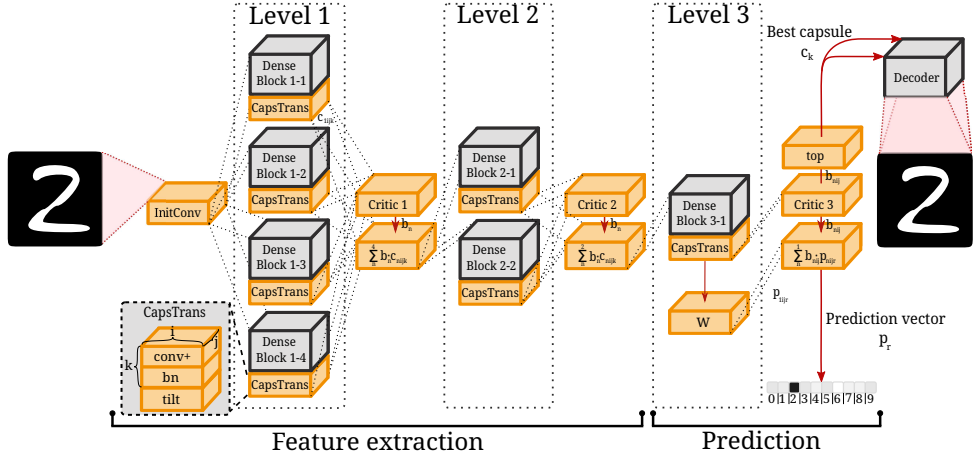}
    \caption{A WCapsNet architecture with four capsule blocks in the first level, two in the second and one in the last level. Each level consists of several independent Dense Blocks followed by a CapsTrans layer, creating the capsule vectors $\mathbf{c}$. The CapsTrans layer consists of the combined \emph{conv+} operation, detailed in Section, and the proposed \emph{tilt} vector non-linearity follow by a batch normalization operation \emph{bn} (see Section \ref{sec:capstrans}). Each level is followed by a critic network assessing the different capsules and prediction weights $\mathbf{b}$ for the capsules. The input for the next level is constructed performing a weighted sum using $\mathbf{b}$. The weights produced by the last critic serve as weighting factors for the prediction vectors and are used to extract the best capsule from the last level.}
    \label{fig:caps_architecture}
    \end{center}
\end{figure*}
We propose a Wasserstein Capsule Network (WCapsNet) using a Wasserstein-critic network to dynamically select features from specialized capsules. We subdivide the network into different levels which are comprised of several capsules. After each level, a critic network assesses the capsules and passes the result of the routing to the next level. This allows the network to dynamically adapt to an input image across multiple levels of depth and abstraction. The levels of WCapsNet can be grouped into two parts, the (i) feature extraction levels, and the (ii) final prediction level, as shown in Figure \ref{fig:caps_architecture}. \newline
Each of the feature extraction levels, consists of $N$ independent capsule blocks $c_{nijk}$, where $i$ and $j$ are the $x$ and $y$ position of a capsule vector with elements $k$, and $n$ is the index of the capsule block. The routing scheme connecting the levels relies on the weighting factors produced by a Wasserstein-critic and performs a weighted sum over the different capsules. For the feature extraction levels, the critic assesses each block $n$ of capsule vectors jointly and assigns a single weight $b_n$ to the whole capsule block $c_{nijk}$, sharing the same weight across all vectors $i$ and $j$ of the 2D map.
The capsule blocks consist of a Dense Block, containing several Dense Layers \cite{DBLP:journals/corr/HuangLW16a}, followed by a capsule transition layer (CapsTrans). The CapsTrans layer consists of a batch normalization operation, a $\mathrm{ReLU}$ activation function and a $1\times 1$ convolution reducing the vector dimension after the Dense Blocks \cite{ioffe2015batch,Hinton_rectifiedlinear}, followed by a vector non-linearity. We propose a vector non-linearity, which is designed to improve the learning behavior for the WCapsNet architecture. The non-linearity tilts the vector into the direction of the strongest vector components and suppresses weak ones. It is presented in more detail in Section \ref{sec:capstrans}.\newline
For the final prediction, in the last level, a critic assigns a separate weight $b_{nij}$ to every capsule vector $c_{nijk}$. Furthermore, a projection matrix $\mathbf{W}$ is used to project the capsule vectors to the one-hot encoded class basis. The weights assigned by the Wasserstein critic are then combined with the projections, using a weighted sum to create the final class prediction of the network.
The capsule vector of the last level with the largest weight is passed to the decoder network (see Fig. \ref{fig:caps_architecture}), to reconstruct the input image. 
The loss of the decoder network consisting of a single fully connected layer and several transposed convolution layers is propagated through the whole network and can therefore modify the capsule vectors to achieve improved reconstruction performance.

\subsection{Capsule transition}\label{sec:capstrans}
The capsule transition layer (CapsTrans), consists of a transition layer applied to the output of the Dense Blocks and a vector non-linearity. The transition layer uses a batch normalization operation, a $\mathrm{ReLU}$ activation function and 1$\times$1 convolution, which we will refer to as a combined conv+ operation (see Fig \ref{fig:caps_architecture}). It produces the vectors $\mathbf{x}_{k}$, where $k$ is the vector dimension. The transition layer is followed by a batch normalization operation and the vector non-linearity, creating the capsule vectors $\mathbf{c}_k$. For the batch normalization before the non-linearity, the parameters are shared among all CapsTrans layers of the level.
In the case of the \emph{squash} non-linearity \cite{NIPS2017_6975},
\begin{equation}
    \mathbf{c}_{k} = \dfrac{||\mathbf{x}_{k}||^2}{1+||\mathbf{x}_{k}||^2}\dfrac{\mathbf{x}_{k}}{||\mathbf{x}_{k}||},
\end{equation}
the function shrinks short vectors close to zero length and long vectors to a value bounded by one.
Since the WCapsNet architecture uses a vector basis projection to recover the class of the input image, we propose an alternative vector non-linearity, that improves the learning behavior of the network. The non-linearity rotates the capsule vectors in the direction of their largest positive components, suppressing weak and attenuating strong elements.  
We use a $\mathrm{softmax}$ function to change the direction of vector $\mathbf{x}$, which we refer to as \emph{tilt} operation,
\begin{equation}
\begin{split}
        \mathbf{c}_k&=\frac{1}{2}\left(\mathbf{1}+\mathrm{softmax}(\mathbf{x}_{k})\right)\odot \mathbf{x}_k,\\
\end{split}
\end{equation}
where $\odot$ indicates an element-wise multiplication. Both non-linearities are empirically compared in Section \ref{sec:experiments}.
\subsection{Wasserstein Objective}
The Wasserstein or Earth-Mover’s distance is an optimal transport distance that is used to approximate distributions. It is defined as:
\begin{equation}
\small
\begin{split}
    W(\mathbb{P}_r,\mathbb{P}_g) &= \inf_{\gamma \in \prod(\mathbb{P}_r,\mathbb{P}_g)} \mathbb{E}_{(x,y)\sim \gamma}\left[\|x-y\|\right],\\
    &= \sup_{\|f\|_L \leq 1} \mathbb{E}_{x\sim \mathbb{P}_r}[f(x)] - \mathbb{E}_{x\sim \mathbb{P}_g}[f(x)]\\ 
\end{split}
\end{equation}
where $\prod (\mathbb{P}_r$,$\mathbb{P}_g)$ denote the set of all joint distributions $\gamma (x,y)$, with the marginals $\mathbb{P}_r$ and $\mathbb{P}_g$. Since finding the supremum is an intractable problem for most cases, an approximate solution is used. Therefore, a neural network representing a Lipschitz function $f(x)$, is trained to maximize the difference between the expectations for samples from both distributions. Approximating the supremum, we obtain $\max_{\|f\|_L \leq 1} \mathbb{E}_{x\sim \mathbb{P}_r}[f(x)] - \mathbb{E}_{x\sim \mathbb{P}_g}[f(x)] $.
In Generative Adverserial Networks (GANs), $f(x)$ is modeled by a neural network called \emph{critic} or discriminator. Here the critic has the task of distinguishing samples from the original distribution of real images, and the inferred distribution of fake images. \newline
To use the Wasserstein distance for a different task such as routing, we first need to find a way to select samples from the distributions we want to distinguish. This requires the occurrence of a specific result or property if samples from at least one of the distributions are present, i.e. a correct or an incorrect classifier prediction. If one can define such a property and therefore distinguish the samples, the corresponding task is defined by the way the critic can influence this property. For our case this means how the routing affects the classification result.  
\subsection{Wasserstein-Routing}\label{sec:ws_routing}
For routing, the task of the Wasserstein-critic $f$ is to identify the best capsules $\mathbf{c}$ for the given input sample $m$. This means that we group the capsules into two distinct distributions, the "good" $p(\mathbf{c}^{(m)})$, and the "bad" capsules $h(\mathbf{c}^{(m)})$. Since we do not want to assign a specific input to a capsule, the distributions $p(\mathbf{c}^{(m)})$ and $h(\mathbf{c}^{(m)})$ are not known a priori. This makes it hard to define a Wasserstein loss for this objective.\newline
However, we can approximate the loss by distinguishing successful routings and failed ones, using the approximate distributions $\tilde{p}(\mathbf{c}^{(m)})$ and $\tilde{h}(\mathbf{c}^{(m)})$. In our approximation, a successful routing is marked by a correct prediction for which the capsules were selected from $\tilde{p}(\mathbf{c}^{(m)})$, and a failed one by a wrong prediction of the network with capsules selected from $\tilde{h}(\mathbf{c}^{(m)})$. \newline
The critic can influence the outcome of the predictions by assigning correct or incorrect weighting factors to the different capsules. If the correct capsules are selected, the prediction is more likely to be correct. This means the critic decides whether a capsules belongs to $p(\mathbf{c}^{(m)})$ or $h(\mathbf{c}^{(m)})$ by assigning a value $f(\mathbf{c}^{(m)})$ to the capsule, which we will refer to as \emph{fitness}. The \emph{fitness} score of capsule block $n$, $f^{(n)}(\mathbf{c}^{(m)})$, relative to the \emph{fitness} of other capsule blocks then reflects the probability of the capsule belonging to the distribution $p(\mathbf{c}^{(m)})$ of the correct capsules. Capsules with a low \emph{fitness} can then be assigned to $h(\mathbf{c}^{(m)})$. 
According to the Wasserstein framework, the critic has to be able to assess single capsules, without comparing the capsule blocks among each other. This constraint limits the amount of available information for the critic, but also comes with the advantage of being less prone to overfitting and having less computational overhead for the routing. 
\subsubsection*{Loss approximation}
The critic network $f$ produces a \emph{fitness} value for each capsule sample $\mathbf{c}^{(m)}$. Over several samples in a mini-batch, the approximate Wasserstein loss $\Tilde{L}_{\mathrm{WS}}$ for a single class, $N$ capsules, $M$ input samples and one critic can be defined as:
\begin{equation}
\small
    \Tilde{\mathrm{L}}_{\mathrm{WS}} = \mathbb{E}_{\mathbf{c}\sim \tilde{h}}[f(\mathbf{c})] - \mathbb{E}_{\mathbf{c}\sim \tilde{p}}[f(\mathbf{c})],\label{eq:approx_loss}
\end{equation}
where $\tilde{p}$ and $\tilde{h}$ are the approximated distributions.\newline
To construct these expectation values, we first need to collect the \emph{fitness} value for each capsule block $\mathbf{c}_n^{(m)}$ and input sample $m$,
\begin{equation}
\small
    a_n^{(m)} = f(\mathbf{c}_n^{(m)}).\label{eq:critic_applied}
\end{equation}
Then a weighting function is applied to the \emph{fitness} values to create the actual capsule weights $b_n$.
The weights $b_n$ are calculated by either applying a $\mathrm{softmax}$ function to $a_n^{(m)}$ along the capsule dimension,
\begin{equation}
\small
    b_n^{(m)} = \mathrm{softmax}(a_n^{(m)}),\label{eq:bn_sm}
\end{equation}
or normalizing the votes according to
\begin{equation}
\small
    b_n^{(m)} = \dfrac{a_n^{(m)}}{\sum_n a_n^{(m)}},\label{eq:bn_norm}
\end{equation}
where $\sum_n b_n^{(m)} = 1$ for both cases.\newline
Based on the weighting factors, we can now determine the approximate values for $\mathbb{E}_{c\sim \tilde{p}}[f(c)]$  and $\mathbb{E}_{c\sim \tilde{h}}[f(c)]$. 
The contributions to the expectation value are the \emph{selected} capsules $F_{\mathrm{s}}$ and the \emph{not selected} capsules $F_{\mathrm{ns}}$, i.e.
\begin{equation}
\small
\begin{split}
    F_{\mathrm{s}}(\mathbf{c}_n^{(m)}) &= \sum_{n=1}^N b_n^{(m)} f(\mathbf{c}_n^{(m)}),\\
    F_{\mathrm{ns}}(\mathbf{c}_n^{(m)}) &= \dfrac{1}{N-1}\sum_{n=1}^N (1-b_n^{(m)}) f(\mathbf{c}_n^{(m)}).\\
\end{split}
\end{equation}{}
The value of $F_{\mathrm{s}}$ should be maximal in case of a correct prediction, while $F_{\mathrm{ns}}$ should be minimal, magnifying the difference for the \emph{fitness} values between correct and incorrect capsules.
Since both the target $\mathbf{t}$ and the prediction vector $\mathbf{p}$ are from $[0,1]$, we can define the correctness of a classification via the cosine distance of the one-hot target vector $\mathbf{t}^{(m)}$ and our prediction vector $\mathbf{p}^{(m)}$:
\begin{equation}
\small
    \cos(\theta)^{(m)} = \dfrac{\mathbf{p}^{(m)}\cdot \mathbf{t}^{(m)}}{||\mathbf{p}^{(m)}||_2 ||\mathbf{t}^{(m)}||_2},
\end{equation}
where $\theta$ is the angle between both vectors. 
We can now select the weight the contributions for the objective, using $\cos(\theta)^{(m)}$. For the correct predictions we assume that the contribution from the selected capsules $F_{\mathrm{s}}$ belongs to the "good" capsules $\Tilde{p}(\mathbf{c}^{(m)})$ and the contribution from the not selected capsules $F_{\mathrm{ns}}$ belongs to the "bad" capsules $\Tilde{h}(\mathbf{c}^{(m)})$. For the case of an incorrect prediction, the only valid assumption is to assign the contribution from the selected capsules $F_{\mathrm{s}}$ to the "bad" capsules $\Tilde{h}(\mathbf{c}^{(m)})$.
The approximate values of $f$ for our distributions $\Tilde{p}(\mathbf{c}^{(m)})$ and $\Tilde{h}(\mathbf{c}^{(m)})$ are:
\begin{equation}
\small
\begin{split}
    f(\Tilde{p}(\mathbf{c}^{(m)})) &= \cos(\theta)^{(m)}  F_{\mathrm{s}}(\mathbf{c}^{(m)}), \\
    f(\Tilde{h}(\mathbf{c}^{(m)})) &= (1-\cos(\theta)^{(m)})F_{\mathrm{s}}(\mathbf{c}^{(m)})+\cos(\theta)^{(m)} F_{\mathrm{ns}}(\mathbf{c}^{(m)}).
\end{split}
\end{equation}
To normalize the loss contributions to be invariant with respect to the amount of correct and incorrect predictions and retrieve expectation values, we calculate normalization factors for a mini-batch of size $M$:
\begin{equation}
\small
\begin{split}
    N_p &= \sum_m^M \cos(\theta)^{(m)}, \\
    N_h &= \sum_m^M (1-\cos(\theta)^{(m)}) .
\end{split}
\end{equation}
Finally, we can construct the expectation values of Eqn. \ref{eq:approx_loss} for a single level, using the approximated distributions,
\begin{equation}
\small
\begin{split}
    \mathbb{E}_{\mathbf{c}\sim \tilde{h}}[f(\mathbf{c})] &=  \dfrac{1}{2 N_h}\sum_{m=1}^M (1-\cos(\theta)^{(m)})F_{\mathrm{s}}(\mathbf{c}^{(m)})\\
    &+\dfrac{1}{2 N_p} \sum_{m=1}^M \cos(\theta)^{(m)} F_{\mathrm{ns}}(\mathbf{c}^{(m)}), \\
    \mathbb{E}_{\mathbf{c}\sim \tilde{p}}[f(\mathbf{c})] & = \dfrac{1}{N_p}\sum_{m=1}^M \cos(\theta)^{(m)}  F_{\mathrm{s}}(\mathbf{c}^{(m)}).
\end{split}
\end{equation}{}
For $\mathbb{E}_{\mathbf{c}\sim \tilde{h}}[f(\mathbf{c})]$ we divide the contributions by a factor of two to balance the expectation losses. This imbalance is rooted in the unknown correct capsule assignment for incorrect predictions. For the critic in the last layer, the $x$ and $y$ position are treated as independent capsules i.e. $\tilde{n} = n\times i\times j$. This leads to $n\times i\times j $ values $a_{\tilde{n}}^{(m)}$ in Eqn. \ref{eq:critic_applied}. 
\subsection{Routing}\label{sec:voting}
The routing relies on the weighting factors $b_n$, produced by the critic network. The input $\tilde{\mathbf{c}}$ for the next level $l+1$ is then calculated by performing a weighted sum over the capsules $\mathbf{c}_{n}^{l}$ of level $l$ :
\begin{equation}
\small
        \tilde{\mathbf{c}}^{l+1} = \sum_n b_n \cdot \mathbf{c}_{n}^{l},
\end{equation}
where $n$ is the capsule, $i$ and $j$ index the location and $k$ the dimensionality of the capsule vector.
\subsection{Prediction}
In the last layer the critic generates weights for each $x$ and $y$ position. This results in a weight vector $b_{nij}$. To create the prediction, we first project the capsule vectors $c_{nijk}$, with the vector elements $k$, to the one-hot basis with elements $r=1\dots N_{\mathrm{Classes}}+1$, using the transformation matrix $W_{kr}$. The weighted sum of all projected vectors then provides the final prediction for the network,
\begin{equation}
\small
     p_{r} = \sum_{nij} b_{nij} \sum_{k} c_{nijk}\cdot W_{kr}.
\end{equation}{}
\subsection{Regularization and loss function}\label{sec:loss_fun}
Since the optimization of a WCapsNet is prone to fall into local optima, we need to employ noise injection and dropout to regularize the training. When selecting capsule blocks the gradient in backpropagation through the selected block is larger than for the other blocks. This leads to better representations within this block, consequently leading to the block being selected more frequently and the routing may collapse. To counteract this issue of selecting always the same capsule we use an additive Gaussian noise from $\mathcal{N}(0,0.5)$ for the \emph{fitness} values. We scale the noise with the maximum of the \emph{fitness} values $\max(a_n^{(m)})$, such that the noise is always in the same order of magnitude as $a_n^{(m)}$. Since using this noise on all values impairs the convergence of the critics, we apply it to 5 \% of the \emph{fitness} values. This provides a good trade-off between sampling the distributions $\tilde{p}$ and $\tilde{h}$ and sufficient convergence of the critics and prevents the routing from collapsing. \newline
To further regularize the training we employ an additional dropout of 0.1 to our weighting factors $b_n$ and a dropout of 0.3 before the projection matrix $\mathbb{W}$.\newline
The training has multiple objectives, therefore the total loss $\mathrm{L}_{tot}$ for the network consists of multiple loss contributions: 
\begin{equation}
\small
    \mathrm{L}_{\mathrm{tot}}=\mathrm{L}_{\mathrm{CE}}+\lambda_{\mathrm{WS}}\cdot \tilde{\mathrm{L}}_{\mathrm{WS}}+\lambda_{\mathrm{R}}\cdot \mathrm{L}_{\mathrm{R}}+\lambda_{\mathrm{WD}} \cdot\mathrm{L}_2,
\end{equation}{}
where $\mathrm{L}_{\mathrm{CE}}$ is the cross entropy loss for the prediction of the network, $\tilde{\mathrm{L}}_{\mathrm{WS}}$ is the Wasserstein loss from the routing process, $\mathrm{L}_{\mathrm{R}}$ is the reconstruction loss for the decoder, and $\mathrm{L}_2$ the regularization loss. The corresponding weighting factors are  $\lambda_{\mathrm{WS}}$, $\lambda_{\mathrm{R}}$ and $\lambda_{\mathrm{WD}}$. We employ the $\mathrm{L}_\mathrm{2}$ weight decay loss to all convolution layers except for the ones used in the CNNs of the Wasserstein critics. 
\section{Network Architecture}\label{sec:network}
The proposed WCapsNet architecture has an exponentially decreasing number of capsules per level, to reflect that complex objects are composed of many different less complex parts. This is also reflected in the dimensions in the capsule vectors. Here the dimension is incremented for the first levels and again decreased for the last level. Decreasing the dimensions in the last level avoids overfitting, since the network needs to generalize to objects. A decoder structure is used to reconstruct the input image, using the best capsules of the last layer as an input.
\subsection{WCapsNet architecture details}
\begin{figure}
    \centering
    \includegraphics[width = 0.6\linewidth]{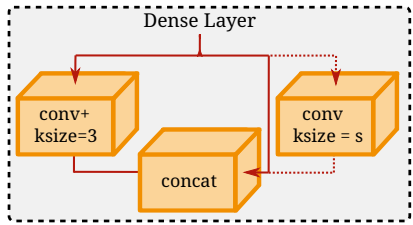}
    \caption{Dense Layer as used in the WCapsNet architecture. For blocks using a stride $s > 1$, the convolution in the combined conv+ (batch normalization, ReLU, convolution) operation uses $s > 1$ and kernel size $ksize$, in the first Dense Layer instead of the transition layer. For this case a shortcut (dotted line) convolution is used to downsample the input for concatenation.}
    \label{fig:dense_layer}
\end{figure}

The WCapsNet uses an initial $3\times 3$ convolution with $24$ channels (InitConv in Figure \ref{fig:caps_architecture}). The result is then passed to the first level of independent Dense Blocks. Contrary to the usual DenseNet architecture as presented in \cite{DBLP:journals/corr/HuangLW16a}, we reduce the spatial dimension of the input within the first layer of the Dense Blocks, rather than in the transition layers.
Since the Dense Blocks need the input of the block for concatenation, we downsample the size of the input using a shortcut convolution layer with a kernel size equal to its stride (see Figure \ref{fig:dense_layer}). This decreases the computational complexity of the model, and does not show significant drops in performance in our experiments.
For the experiments we use a 4 level WCapsNet, with $N$ = 16-8-4-2 capsule blocks. The number of Dense Layers per capsule was fixed to $n_D=6$ for all networks. Other parameters used in the WCapsNets are shown in Table \ref{tab:architecture_1}.
\begin{table}[h]
\centering
    \caption{Parametrization of the WCapsNet architectures used for the dataset MNIST, SVHN and CIFAR-10/100. The growth rate for the Dense Blocks is denoted as $g$, stride is the downsampling parameter of the Dense Blocks, and the vector dimensionality of the CapsTrans is referred to as $k$. params. denotes the number of parameters.}
    \label{tab:architecture_1}
    \begin{tabular}{l|c|c|c||c|c|c}
    \hline
    &\multicolumn{3}{c|}{CIFAR-10 / SVHN}&\multicolumn{3}{|c}{CIFAR-100}\\ 
    \hline
      &$g$& $k$&stride &$g$& $k$&stride \\
    \hline
       Level 1 &8&16&2&8&16&2\\
    \hline
       Level 2 &8&32&1&8&32&1\\
    \hline
       Level 3 &8&64&2&8&64&2\\
    \hline
       Level 4 &8&8&1&8&24&1\\
    \hline
    \hline
    Classifier params.&\multicolumn{3}{|c|}{697 k}&\multicolumn{3}{|c|}{701 k} \\ 
    \hline
    Critic params.&\multicolumn{3}{|c|}{210 k}&\multicolumn{3}{|c|}{213 k} \\ 
    \hline
    Decoder params.&\multicolumn{3}{|c|}{43 k}&\multicolumn{3}{|c|}{76 k} \\ 
    \hline
    Total params.&\multicolumn{3}{|c|}{950 k}&\multicolumn{3}{|c|}{990 k} \\ 
    \hline
    \end{tabular}
\end{table}
\subsection{Critic CNN}
Since the critic in the last level needs to provide a separate weight for each individual capsule vector, whereas the other critics do a block wise weighting, two different architectures are implemented.\newline
(i) The feature extraction critic architecture is used for all levels except for the last one.
It consists of $3\times 3$ convolutions with a stride of $s=2$, followed by a $\mathrm{ReLU}$ activation function and a dropout layer with a dropout rate of $r=0.3$. We increase the number of channels per layer as the height and width decreases.
For layer $j$ the number of used channels is $n_{\mathrm{ch}}= j \cdot k_{critic}$. In our experiments we use $k_{critic}=32$ for the convolutions. The number of layers of each critic depends on the size of the input. This means layers are added until the size is downsampled to one and we receive a single value as our output.\newline
(ii) The critic for the last level has the same structure, but uses 4 layers of $1\times 1$ convolutions with a stride of 1, therefore the output has the same size as the input, providing $height\times width$ \emph{fitness} values. 
To limit the critic outputs and restrict the values to the interval $[0,1]$, we apply a batch normalization followed by a $\mathrm{sigmoid}$ function to all output values. The convolution kernels of the critic CNNs use spectral normalization on the weights, ensuring the Lipschitz criterion of $f$ \cite{miyato2018spectral}. 
The gradient from the critic to the capsules is stopped during training, so the critic cannot modify the capsule blocks. 
\subsection{Decoder and reconstruction loss}\label{sec:decoder}
The decoder network has the task of reconstructing the input based on the selected capsule vector. The gradients from the reconstruction are propagated through the whole network and can therefore influence the capsule vectors, leading to better representations. For our experiments we use the best vector of the last level in the decoder structure. We add the vector position of the extracted capsule vector by concatenating the vector with $x$ and $y$ coordinates, normalized to [-1,1]. The decoder structure consists of one fully connected layer, creating a 2D patch a quarter of the original input size large. Now we apply two transposed convolution layers with a stride of two to create the decoder output. The convolution layers use 32 for the first and 64 channels for the second convolution. Each of the convolution layers employs a batch normalization operation and a $\mathrm{ReLU}$ activation before the convolution.\newline
We use a Mean Squared Error (MSE) loss to train the network to reconstruct the input image based on the best capsule vector.
\section{Experiments}\label{sec:experiments}
We conduct several experiments evaluating the WCapsNet architecture. We perform ablation studies for the proposed routing scheme and the \emph{tilt} vector non-linearity. Therefore, we train WCapsNet on a image classification task using several standard image datasets. Furthermore, we investigate the voting in more detail for the MNIST dataset. We analyze the capsule weighting factors $b_n$ (see Equation \ref{eq:bn_sm} and \ref{eq:bn_norm}) for different classes across multiple levels. The networks use the parametrization of Table \ref{tab:architecture_1}.
\subsection{Datasets and training setup}\label{sec:dataset_img}
We select 5 benchmark datasets to evaluate our WCapsNet architecture.
\begin{itemize}
    \item \textbf{MNIST}  (\cite{LeCun2010}): A set of centered 28$\times$28 handwritten digits from 0-9 in black and white. It consists of 60000 training samples and 10000 test samples. The dataset was normalized to the interval $[0,1]$. We use a training/validation split of 50000/10000 images.
    \item \textbf{CIFAR-10} (\cite{Krizhevsky}): The CIFAR datasets are RGB image datasets displaying real world objects at a resolution of 32$\times$32. The CIFAR-10 dataset includes ten different types of objects. It consists of 50000 training and 10000 test samples. We adopt a standard data augmentation scheme including, standardization, mirroring and shifting of the images. We use a training/validation split of 45000/5000 images.
    \item \textbf{CIFAR-100} (\cite{Krizhevsky}): This dataset has the same specifications as CIFAR-10, but consists of 100 classes of objects. For training we use the same data augmentation as for CIFAR-10. We use a training/validation split of 45000/5000.
    \item \textbf{SVHN} (\cite{SVHN}): This RGB image dataset consists of house numbers taken from Google Street View, with a single digit to classify. It consists of 73257 training samples and 26032 test samples at a resolution of 32$\times$32, the dataset was normalized to the interval $[0,1]$. We use a training/validation split of 63257/10000 images.
\end{itemize}
\subsection{Training settings}
Since the architecture uses Dense Blocks, we use the training setup of DenseNet as presented in \cite{DBLP:journals/corr/HuangLW16a}. We use a stochastic gradient descent optimizer with a Nesterov momentum of 0.9, using a batch size of 64. For the CIFAR datasets, we use a base learning rate of 0.1 and decay the learning rate after 150, 200 and 250 epochs by a factor of 0.1. The dropout rate in the Dense Blocks is set to zero. For MNIST and SVHN we train the network for a maximum of 40 epochs and decay the learning rate after 20 and 30 epochs by a factor of 0.1. 
We used a weight decay scaling factor of $\lambda_{\mathrm{WD}}=10^{-4}$ and a scaling factor of $\lambda_{\mathrm{WS}}=0.2$ for the Wasserstein loss and $\lambda_{\mathrm{R}}=0.1$ for the reconstruction loss. 
\subsection{Ablation studies}\label{sec:ablation_studies}
To verify our WCapsNet architecture, we perform several ablation studies. We investigate different variants of the routing scheme, different weighting functions and vary the vector non-linearity of the network. All our results were generated using early stopping using the train/validation splits mentioned in Section \ref{sec:dataset_img}.
\paragraph{Weighting functions}
We compare the results of the network using either Eqn. \ref{eq:bn_sm} or \ref{eq:bn_norm} as a weighting function for the routing weights.
\begin{table}[h]
    \centering
        \caption{Comparision of the results for different weighting function for the capsule routing.} 
    \begin{tabular}{l||c|c|c|c}
        \hline
        Variant&CIFAR-10&CIFAR-100&SVHN&MNIST \\
        \hline
        $\mathrm{softmax}$&\textbf{93.43} \%&\textbf{70.39} \%&\textbf{96.46}\%&\textbf{99.68}\%\\
        \hline
        Normalized&93.04\%&69.75 \%&96.33 \%&99.58\%\\
        \hline
    \end{tabular}
    \label{tab:results_weighting}
\end{table}
The results in Table \ref{tab:results_weighting} show that the $\mathrm{softmax}$ weighting function achieves slightly better results than simple normalization of the votes.
\paragraph{Investigation of different vector non-linearities}
We compare the \emph{tilt} vector non-linearity to the \emph{squash} non-linearity, using a $\mathrm{softmax}$ weighting function for the routing. The first variant represents the baseline only using the \emph{squash} non-linearity. For the second variant we use the \emph{tilt} non-linearity. 
\begin{table}[h]
    \centering
        \caption{Comparision of the results for different vector non-linearities.} 
    \begin{tabular}{l||c|c|c|c}
        \hline
        Variant&CIFAR-10&CIFAR-100&SVHN&MNIST \\
        \hline
        \emph{squash}&92.91 \%&64.42\%&\textbf{96.51}\%&99.64\%\\
        \hline
        \emph{tilt}&\textbf{93.43} \%&\textbf{70.39} \%&96.46\%&\textbf{99.68}\%\\
        \hline
    \end{tabular}
    \label{tab:results_nlin}
\end{table}
The results in Table \ref{tab:results_nlin} show, that the \emph{tilt} non-linearity outperforms the \emph{squash} non-linearity especially for more complex tasks as CIFAR-100. This indicates that the \emph{tilt} non-linearity improves the network behavior.
\paragraph{Comparing different routing variations}
We compare different variants of training the routing networks. The first variant does not stop the gradient before the weighting factors $b_n$, and therefore uses the cross-entropy $\mathrm{CE}$ and the Wasserstein loss $\mathrm{WS}$ to train the critic networks. The second variant stops the gradient from the cross-entropy loss, and is only trained using the Wasserstein loss. The third variant does not use a Wasserstein objective to train the routing networks, this means that the routing weights are adjusted using only the cross-entropy loss. The fourth variant uses random routing weights drawn form a uniform distribution which is then normalized such that $\sum_n b_n =1$. The last variant uses a uniform weight distribution which means that all weights are set to $b_{i} = \frac{1}{\mathrm{N}}$.
\begin{table}[h]
    \centering
        \caption{Comparision of the results for different routing variants.} 
    \begin{tabular}{l||c|c|c|c}
        \hline
        Variant&CIFAR-10&CIFAR-100&SVHN&MNIST \\
        \hline
        $\mathrm{WS+CE}$&\textbf{93.43} \%&\textbf{70.39} \%&96.46\%&\textbf{99.68}\%\\
        \hline
        $\mathrm{WS}$&92.54 \%&69.10 \%&\textbf{96.56}\%&99.60\%\\
        \hline
        $\mathrm{CE}$&93.05 \%&70.30 \%&96.39 \%&99.65 \%\\
        \hline
        $\mathrm{Random}$&91.90 \%&66.81 \%&96.20 \%&99.49\%\\
        \hline
        $\mathrm{Uniform}$&93.00 \%&70.14 \%&96.37\%&99.64 \%\\
        \hline
    \end{tabular}
    \label{tab:results_routing}
\end{table}
As we can see in Table \ref{tab:results_routing}, the variant using both the cross-entropy and the Wasserstein loss performs the best for most datasets. The Wasserstein loss alone only works well for very simple dataset and does not perform well for the more complex CIFAR datasets, which is not surprising since the objective does not optimize the classification result, but rather takes it as given.

\subsection{Image Classification}\label{sec:image_classification}
In Table \ref{tab:results} we compare WCapsNets, to other capsule architectures using the best results of our experiments. 
\begin{table*}
    \centering
        \caption{Results in terms of test accuracy and number of parameters. Parameters are given for the CIFAR-10 dataset and architecture.*images are resized to 64$\times$64 pixels, in contrast to all other papers using the original size of 32$\times$32 pixels.} 
    \begin{tabular}{l|c|c|c||c||c||c}
        \hline
        Architecture&Method&Params&CIFAR-10&CIFAR-100&SVHN&MNIST \\
        \hline
        &DenseNet 250 \cite{DBLP:journals/corr/HuangLW16a},\cite{DBLP:journals/corr/abs-1904-09546},\cite{ferrari2018computer} & 15.3 M & 96.40 \%&82.40 \%&98.41 \%& -\\
        DNN & ResNet 1001 (pre-act.) \cite{10.1007/978-3-319-46493-0_38} & 10.2 M &95.38 \%&77.30 \% &-& -\\
        &VGG 19  \cite{simonyan2014very,tetko2019artificial}& 20 M & 93.66 \%& 73.26 \% &-\\
        \hline
        &CapsNets \cite{NIPS2017_6975} & 8.2M &  70.01 \%&- & 91.71\%& 99.62\%  \\
        Capsule Network&DeepCaps \cite{DBLP:journals/corr/abs-1904-09546} &8.5M &91.01\% *&-&\textbf{97.16\%} *&\textbf{99.72\%}\\
        &Self-Routing CapsNet \cite{hahn2019self} &3.2M &92.14\% &-&96.88\%&-\\
        &WCapsNet (this work)&950k&\textbf{93.43\%}&\textbf{70.39}\%&96.56\%&99.68\%\\
        \hline
    \end{tabular}
    \label{tab:results}
\end{table*}
The results show, that WCapsNets can substantially outperform other capsule approaches on CIFAR-10, while having a fraction of the parameters. The classification performance of the WCapsNet on CIFAR-100 is lower compared to large state-of-the art CNN architectures, but comes close in performance to older DNN architectures like VGG-19.
\subsection{Network evaluation}
We evaluate the routing weights $b_n$ assigned by the critics for each level of the network. The distribution of weighting factors shows the degree and type of specialization of each capsule. The evaluation of the prediction vectors provides information about the assignability of a feature to a specific class, and therefore indicates the complexity of the features in each level. 
\begin{figure*}[h]
\begin{minipage}[t]{0.89\textwidth}
        \resizebox {\linewidth} {!} {
            \input{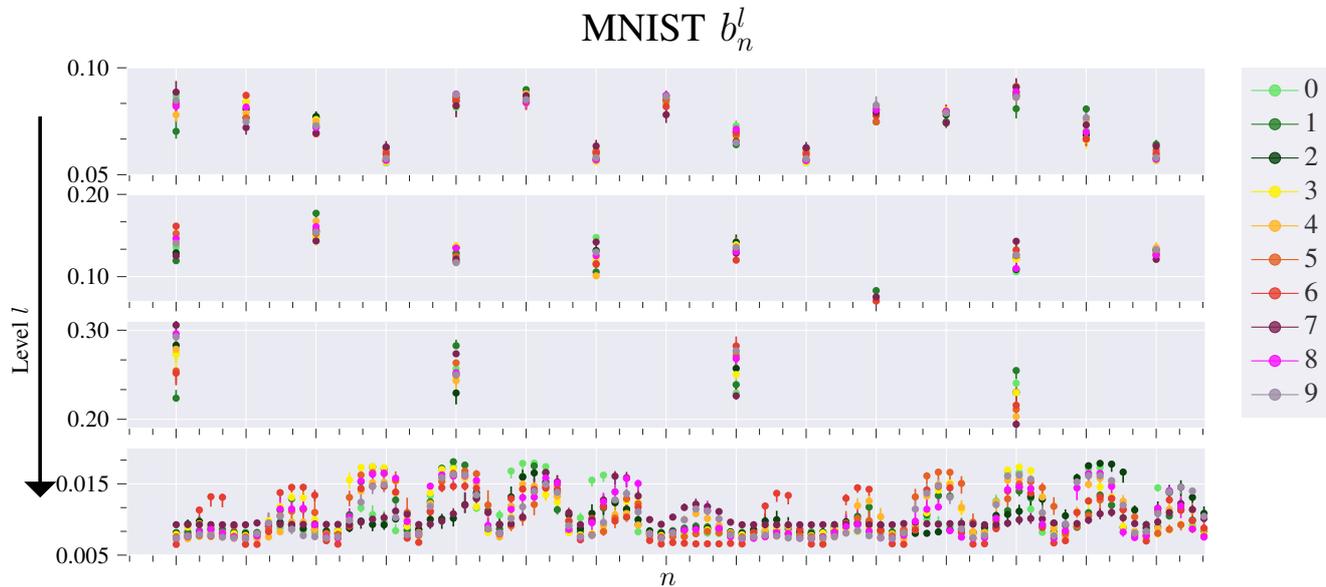}
        }
\end{minipage}
\begin{minipage}[t]{0.1\textwidth}
        \resizebox {\linewidth} {!} {
\begin{tikzpicture}

\definecolor{color0}{rgb}{0.917647058823529,0.917647058823529,0.949019607843137}
\definecolor{color1}{rgb}{0.403921568627451,0.898039215686275,0.407843137254902}
\definecolor{color2}{rgb}{0.145098039215686,0.498039215686275,0.152941176470588}
\definecolor{color3}{rgb}{0.0313725490196078,0.258823529411765,0.0509803921568627}
\definecolor{color4}{rgb}{1,0.941176470588235,0}
\definecolor{color5}{rgb}{1,0.713725490196078,0.168627450980392}
\definecolor{color6}{rgb}{0.898039215686275,0.380392156862745,0.141176470588235}
\definecolor{color7}{rgb}{0.898039215686275,0.243137254901961,0.188235294117647}
\definecolor{color8}{rgb}{0.498039215686275,0.137254901960784,0.325490196078431}
\definecolor{color9}{rgb}{0.976470588235294,0.0666666666666667,1}
\definecolor{color10}{rgb}{0.623529411764706,0.549019607843137,0.650980392156863}

\node (SHIFT) at (-1, -6.6) {};

\begin{axis}[%
hide axis,
width=1\textwidth,
height=1\textwidth,
xmin=0,
xmax=0.5\textwidth,
ymin=0,
ymax=0.5\textwidth,
legend cell align={left},
legend style={fill opacity=0.8, draw opacity=1, text opacity=1, at={(0,0)}, draw=none, fill=color0}
]
\addlegendimage{color1,mark=*}
\addlegendentry{ 0}
\addlegendimage{color2,mark=*}
\addlegendentry{ 1}
\addlegendimage{color3,mark=*}
\addlegendentry{ 2}
\addlegendimage{color4,mark=*}
\addlegendentry{ 3}
\addlegendimage{color5,mark=*}
\addlegendentry{ 4}
\addlegendimage{color6,mark=*}
\addlegendentry{ 5}
\addlegendimage{color7,mark=*}
\addlegendentry{ 6}
\addlegendimage{color8,mark=*}
\addlegendentry{ 7}
\addlegendimage{color9,mark=*}
\addlegendentry{ 8}
\addlegendimage{color10,mark=*}
\addlegendentry{ 9}
\end{axis}
\end{tikzpicture}
        }
\end{minipage}
    \caption{Results for the evaluation of the routing weights $b_n$ (left) for the MNIST Dataset. The colored dots show the mean weight given to the specific class.}\label{fig:eval_MNIST}
\end{figure*}
The results of MNIST shown in Fig. \ref{fig:eval_MNIST} for the average per class weighting factors $b_n$ show that the network does specialize the capsules to specific classes. Capsules in deeper levels are more likely to specialize to a larger degree, whereas in the first levels only slight changes in the weighting are present. This supports our assumption that capsules in the first levels represent parts of objects which occur across multiple classes. For the third level, which shows substantial specialization, we see that capsule block two is specialized to detect the number one, whereas capsule three has a large weighting factor if a five or nine is present. The routing weights of the last level contain a periodicity which is related to the $x$ and $y$ positions, but also contains a lot of inter class variation between the weighting factors for the same position. However, the specialization of the capsules is not as large as one might expect. This might be caused by the optimization process, since high routing weight specialization can cause temporary drops in performance during training. 
\section{Conclusion and Outlook}
We propose a capsule network architecture (WCapsNet), which can dynamically adapt to the input image. The dynamic routing procedure relies on a neural network, called \emph{critic}, that is trained with an approximate Wasserstein objective. We propose an approximation scheme for the Wasserstein loss suitable for solving the task of routing. Furthermore, we propose a direction dependant vector non-linearity suited for the proposed capsule architecture.
WCapsNets offers a new and scale-able approach for image classification and improves the interpretability of classification results, by offering a possibility to analyze capsule weights at multiple levels.
The classification results show that WCapsNets are able to achieve less than 6.6\% of error on CIFAR-10 outperforming other capsule approaches. Furthermore, WCapsNets are able to achieve good performance on CIFAR-100, which was not feasible with previous capsule architectures that relied on vector length based classification rather than vector projections. We analyze the routing weights for the proposed Wasserstein-routing and visualize the capsule specializations after each level.
For future research we would like to explore different methods for training the WCapsNet architecture to achieve a higher degree of specialization within the capsules, and apply WCapsNets to a supervised segmentation tasks leveraging its benefits in more realistic applications.

{\small
\bibliographystyle{IEEEtran}
\bibliography{bibfile}
}

\end{document}